\documentclass{article}

 \usepackage[preprint]{neurips_2026}

\usepackage[utf8]{inputenc} 
\usepackage[T1]{fontenc}    
\usepackage{hyperref}       
\usepackage{url}            
\usepackage{booktabs}       
\usepackage{amsfonts}       
\usepackage{nicefrac}       
\usepackage{microtype}      
\usepackage{xcolor}         
\usepackage{graphicx}

\usepackage[T1]{fontenc}
\usepackage{graphicx}
\usepackage{booktabs}
\usepackage{array}
\usepackage{tabularx}
\usepackage{multirow}
\usepackage{rotating}
\usepackage{hyperref}
\usepackage{color}
\usepackage[normalem]{ulem}
\usepackage{amsmath}
\usepackage{microtype}
\usepackage{listings}
\usepackage{caption}
\usepackage{subcaption}
\usepackage{enumitem}

\title{Nepali Legal Expertise through Generative and Extractive Pre-trained Transformers (NepLEGiT)}

\author{%
  Ranjit Raut \\
  Department of Artificial Intelligence \\
  Kathmandu University \\
  Dhulikhel, Nepal \\
  \texttt{rautranjit916@gmail.com} \\
  \And
  Tishya Dhakal \\
  Department of Artificial Intelligence \\
  Kathmandu University \\
  Dhulikhel, Nepal \\
  \texttt{tishys57@gmail.com} \\
  \And
  Aaryan Shakya \\
  Department of Artificial Intelligence \\
  Kathmandu University \\
  Dhulikhel, Nepal \\
  \texttt{nayranewar33@gmail.com} \\
  \And
  Bhabuk Thapa \\
  Department of Artificial Intelligence \\
  Kathmandu University \\
  Dhulikhel, Nepal \\
  \texttt{vhabukthapa@gmail.com} \\
  \And
  Prasiddha Koirala \\
  Department of Artificial Intelligence \\
  Kathmandu University \\
  Dhulikhel, Nepal \\
  \texttt{prasiddhaf23@gmail.com} \\
  \And
  Bal Krishna Bal \\
  Department of Computer Science and Engineering \\
  Kathmandu University \\
  Dhulikhel, Nepal \\
  \texttt{bal@ku.edu.np} \\
}

\begin{document}
\pagestyle{plain}
\maketitle
\begin{abstract}
The complexity of legal language and limited accessibility to legal information pose significant challenges to justice delivery in Nepal. Traditional legal services remain inaccessible to many citizens due to language barriers, information fragmentation, and a critical shortage of legal expertise, particularly in rural areas. We present \textbf{NepLEGiT} (\textbf{N}epali \textbf{L}egal \textbf{E}xpertise through \textbf{G}enerative and \textbf{E}xtractive Pre-tra\textbf{i}ned \textbf{T}ransformers), a specialized small language model (SLM) designed to democratize legal knowledge and enhance legal-service delivery in Nepal. We pre-train a decoder-based GPT-2 SLM from scratch on a curated corpus of $\sim$4 million tokens of Nepali legal text, covering constitutional law, civil and criminal codes, and administrative regulations. The model comprises $\sim$30 million parameters in a 6-layer, 6-head, 384-dimensional transformer trained with warmup cosine-decay scheduling, gradient accumulation, and mixed-precision arithmetic. On a held-out validation split, NepLEGiT attains a cross-entropy loss of 0.5684, a perplexity of 1.8, and a next-token prediction accuracy of 82.9\%. We further evaluate continual masked-language-model pre-training of mBERT and MuRIL on the same corpus; mBERT achieves a perplexity of 2.35 (eval loss 0.8565), outperforming MuRIL (perplexity 6.07, eval loss 1.8026), providing a strong encoder baseline complementary to NepLEGiT's generative orientation. 


\end{abstract}
\section{Introduction}
\label{sec:intro}

The digitization of legal knowledge has catalyzed a wave of legal AI systems tailored to document analysis, question answering, and compliance assistance. Yet the overwhelming majority of these systems target English-language, common-law jurisdictions, leaving the legal systems of the Global South largely unaddressed. Nepal exemplifies this gap: its legal framework blends elements of civil law, common law, and indigenous customary practices, written primarily in formal Nepali suffused with Sanskrit-derived terminology that places it beyond the effective reach of general-purpose multilingual models.

The judiciary of Nepal handles an ever-growing caseload under conditions of sparse digitization and acute shortages of qualified legal professionals in rural districts. Citizens who cannot afford lawyers often navigate the legal system without assistance, compounding existing inequalities. Meanwhile, the country's Digital Nepal Framework articulates a mandate for e-governance and digitally empowered citizenship, a mandate that AI-powered legal tools could help fulfill.

The challenges of Nepal's legal accessibility include:
(i)~Language barriers: legal documents use formal Nepali with Sanskrit-derived terminology that hinders comprehension for average citizens;
(ii)~Limited legal expertise: legal information is scattered across gazettes, court decisions, and orders, with severe shortages of qualified professionals in remote districts; (iii)~Technological gap: existing legal AI targets Western (common/civil law) systems and English, ignoring Nepal's mixed jurisprudence; and
(iv)~Compliance complexity: rapidly changing regulations cause inadvertent violations among businesses and citizens. We address this need with a 10M-token dataset sourced from the Nepal Law Commission, with a 4M-token clean training split, a GPT-2 decoder-based language model pre-trained from scratch on a curated Nepali legal corpus, and continual pre-training masked-language-model mBERT and MuRIL.



\section{Related Work}
\label{sec:related}


Domain-specific pre-training on legal corpora has consistently outperformed general-purpose models on legal NLP tasks. LegalBERT \cite{zheng2021} demonstrated this for English by continuing BERT pre-training on contracts, court opinions, and statutory texts, yielding improvements on document classification and named-entity recognition. LexGLUE \cite{chalkidis2022} consolidated evaluation for English legal NLP across classification, NER, QA, and entailment tasks, establishing that in-domain pre-training is the most robust single intervention for legal model performance. Harvey AI and Legal Robot extend this to GPT-4-powered generative settings, though both remain restricted to common-law, English-language contexts.


mBERT and XLM-R \cite{conneau2020} provide broad multilingual coverage but underperform in highly specialized domains such as law, where vocabulary and syntactic conventions diverge sharply from general-domain training data. MuRIL \cite{khanuja2021} demonstrated that language-specific fine-tuning substantially benefits Indic languages including Nepali, supporting our choice to pre-train on a Nepali-only corpus. In our experiments, continual masked-language-model pre-training of mBERT on the Nepali legal corpus yields a perplexity of 2.35, outperforming MuRIL (perplexity 6.07) on this domain --- an important empirical finding suggesting that broad multilingual coverage can be more amenable to legal domain adaptation than Indic-focused pre-training when the target language is Nepali. No prior work, however, has addressed Nepali legal NLP at the pre-training stage.


JEC-QA \cite{zhong2020} provides a landmark dataset for Chinese legal question answering and demonstrates that non-Western legal systems with distinct jurisdictions, unique citation conventions, and non-Latin scripts require bespoke solutions. COLIEE \cite{rabelo2019} advances case-retrieval for common law but does not generalize beyond English. NepLEGiT is positioned analogously to JEC-QA but for the Nepali legal and linguistic context.


\citet{eldan2023} showed that small models trained on carefully curated domain-specific data can match or exceed much larger general-purpose models on targeted tasks. This motivates our choice to pre-train a $\sim$30M-parameter model rather than fine-tune a multi-billion-parameter LLM: parameter efficiency, fast inference, and deployability under resource constraints are essential for legal AI in a developing-country context.


The transformer \cite{vaswani2017} introduced self-attention as the foundation for all contemporary language models. BERT \cite{devlin2019} and RoBERTa \cite{liu2019} popularized bidirectional encoder pre-training; GPT and its successors \cite{radford2018,radford2019,brown2020} established autoregressive decoder pre-training as the dominant paradigm for text generation. BART \cite{lewis2020} and T5 \cite{raffel2020} combined both into encoder--decoder architectures suited to seq2seq tasks. Our architecture evaluation (Section~\ref{sec:arch}) leads us to select GPT-2 for NepLEGiT's generative orientation.

\section{Methodology}

\subsection{Dataset}
\label{sec:corpus}

We made a Nepali legal corpus from the Nepal Law Commission: Constitution, civil and criminal codes, and all major statutes.


\begin{table}[htbp]
    \centering
    \small
    \caption{Nepali Legal Corpus Statistics}
    \label{tab:corpus}
    \begin{tabular}{@{}lr@{}}
        \toprule
        Attribute & Value \\
        \midrule
        Total documents collected & $\sim$1,000 \\
        Total pages processed     & $\sim$1,000,000 \\
        Total tokens extracted    & $\sim$10,000,000 \\
        Time period covered       & 1816--2025 \\
        Language distribution     & Nepali 95\%, English 3\%, Mixed 2\% \\
        \midrule
        Training split (90\%)     & $\sim$4,000,000 tokens \\
        Validation split (10\%)   & $\sim$400,000 tokens \\
        \bottomrule
    \end{tabular}
\end{table}

\subsection{Models}
\label{sec:arch}


Bidirectional encoding excels at understanding tasks: classification, NER, extractive QA, but the absence of a causal decoder renders it ill-suited for open-ended legal text generation, the primary use case of NepLEGiT. We select mBERT and MuRIL for NepLEGiT. Sequence-to-sequence architectures offer versatility across comprehension and generation and are well-suited to summarization and generative QA. Their dual-stack design doubles computational cost and requires substantially larger datasets to converge effectively. Autoregressive decoders produce coherent, context-consistent outputs over extended sequences. The causal language-modelling objective is straightforward to optimize, scales predictably, and directly serves NepLEGiT's primary use cases: legal document drafting, explanation generation, and conversational legal guidance. We select decoder-based GPT-2 for NepLEGiT. We evaluated three transformer families against the requirements of Nepali legal text generation.

\begin{figure}[htbp]
    \centering
    \includegraphics[width=0.9\linewidth]{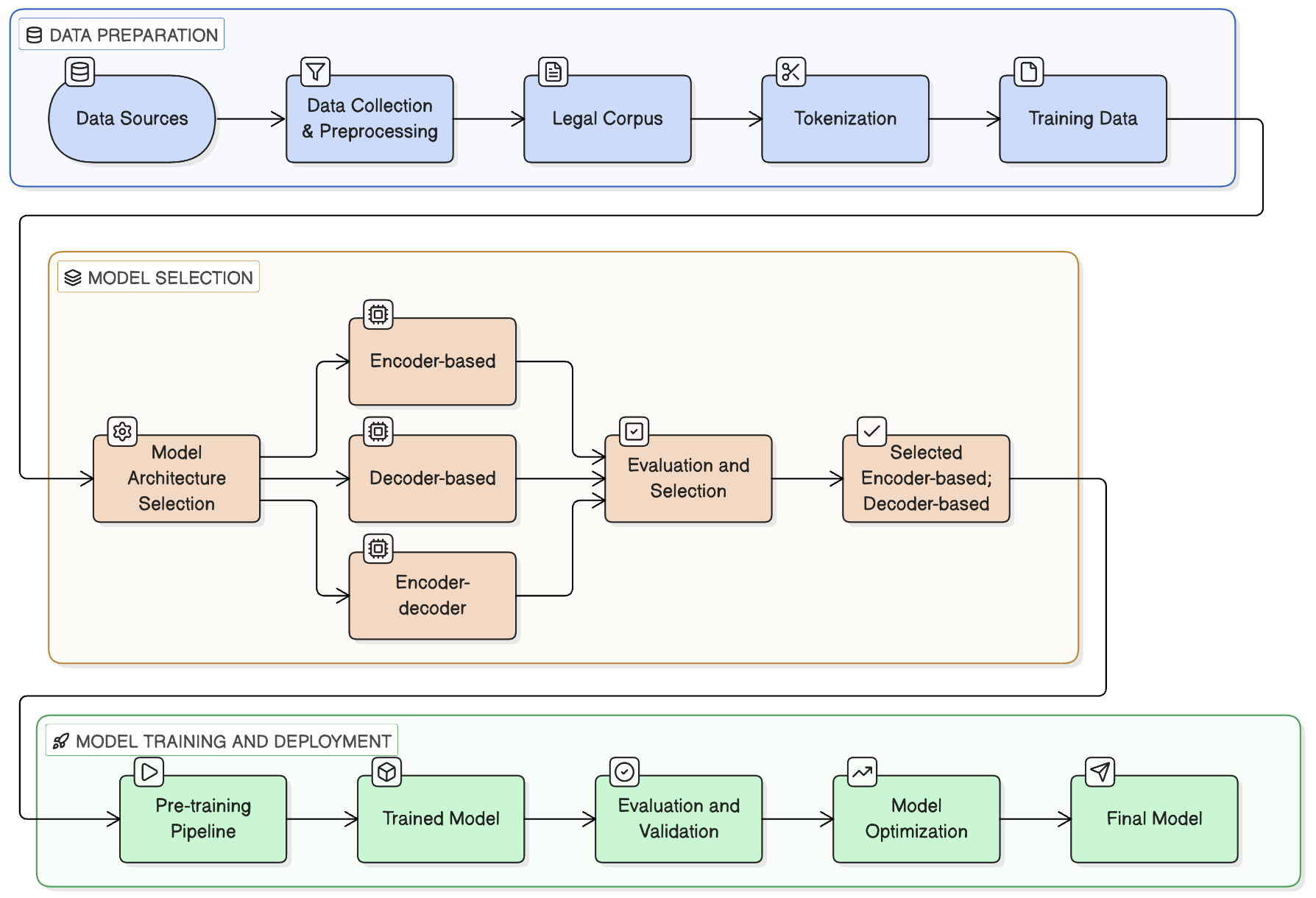}
    \caption{System Architecture.}
    \label{fig:system_architecture}
\end{figure}

We adopt the GPT-2 BPE tokenizer (Tiktoken library, vocabulary size~50,257) for its robustness to Unicode characters in NepLEGiT's GPT-2 architecture. After NFC normalization, it handles Nepali subword segmentation acceptably; average legal document length is $\sim$2,500 tokens. The preprocessed corpus was serialized to binary NumPy \texttt{memmap} format, enabling efficient random-access I/O without loading the full dataset into RAM. In the context of mBERT and MuRIL, we used their own dedicated tokenizer config and vocabulary hosted on Hugging Face, which we loaded using the standard AutoTokenizer classes.


\begin{figure}[htbp]
    \centering
    \includegraphics[width=0.9\linewidth]{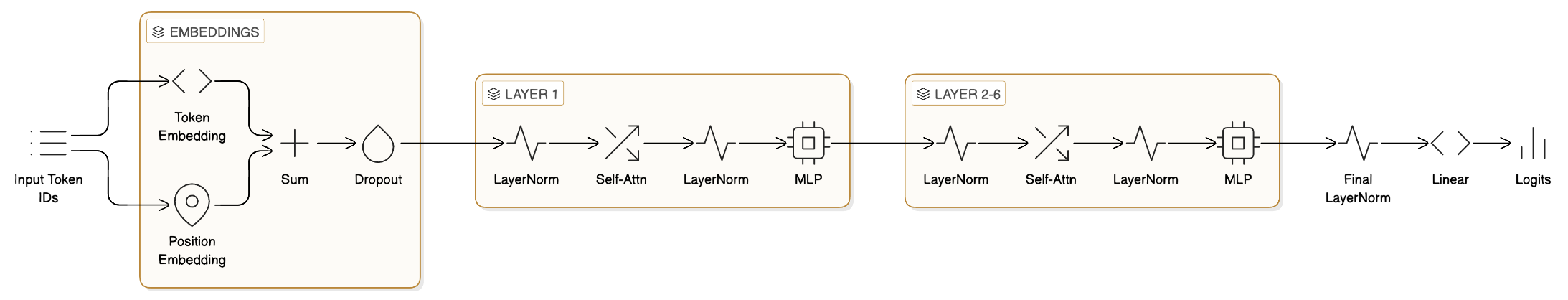}
    \caption{GPT-2 Model Architecture.}
    \label{fig:model}
\end{figure}





NepLEGiT implements a standard causal transformer decoder following \citet{vaswani2017} with the modifications of \citet{radford2019}. Table~\ref{tab:config} summarizes the configuration.

\begin{table}[htbp]
    \centering
    \small
    \caption{NepLEGiT GPT-2 Model Configuration}
    \label{tab:config}
    \begin{tabular}{@{}lll@{}}
        \toprule
        Parameter & Value & Description \\
        \midrule
        \texttt{vocab\_size}  & 50,257    & GPT-2 BPE vocabulary \\
        \texttt{block\_size}  & 128       & Context window (tokens) \\
        \texttt{n\_layer}     & 6         & Transformer blocks \\
        \texttt{n\_head}      & 6         & Attention heads per block \\
        \texttt{n\_embd}      & 384       & Embedding dimension \\
        \texttt{dropout}      & 0.1       & Regularization dropout \\
        \texttt{bias}         & True      & Bias in linear layers \\
        Total parameters      & $\sim$30M & Trainable parameters \\
        Model size (disk)     & $\sim$240\,MB & Saved checkpoint \\
        \bottomrule
    \end{tabular}
\end{table}


Layer normalization is applied \emph{before} the attention and MLP sub-layers (rather than after), stabilizing gradient flow in deep networks \cite{brown2020}. Feed-forward sub-layers use GELU with a $4\times$ hidden-dimension expansion ($4 \times 384 = 1{,}536$). The output projection matrix is tied to the input token embedding, reducing parameters and enforcing representational consistency between input and output spaces. Scaled dot-product self-attention is masked to prevent attention to future positions. Flash attention is enabled where hardware supports it. Summed with token embeddings at the input layer. Linear layers use $\mathcal{N}(0, 0.02)$; residual projection layers use $\mathcal{N}(0,\, 0.02/\!\sqrt{2\,n_{\text{layer}}})$ following \citet{radford2019}.






\subsection{Training}
\label{sec:training}



\begin{table}[htbp]
    \centering
    \small
    \caption{Training Hyperparameters}
    \label{tab:hparam}
    \begin{tabular}{@{}ll@{}}
        \toprule
        Hyperparameter & Value \\
        \midrule
        Peak learning rate     & $1\times10^{-4}$  \\
        Min.\ learning rate    & $5\times10^{-5}$  \\
        Warmup steps           & 1,000             \\
        Total iterations       & 100,000           \\
        Micro-batch size       & 32                \\
        Gradient accum.\ steps & 32                \\
        Gradient clip norm     & 0.5               \\
        Weight decay           & 0.1               \\
        $\beta_1,\;\beta_2$    & 0.9,\ 0.95        \\
        $\varepsilon$          & $10^{-9}$         \\
        Precision              & bfloat16/float16  \\
        Eval.\ interval        & 500 steps         \\
        \bottomrule
    \end{tabular}
\end{table}


We use AdamW with decoupled weight decay \cite{loshchilov2018}. The higher $\beta_2 = 0.95$ (versus the conventional 0.999) produces smoother second-moment estimates that benefit sparse, domain-specific vocabulary updates. We implement a two-phase schedule via PyTorch's \texttt{SequentialLR}.



Phase 1 -- Linear warmup (steps 0--1,000)
\[
    \eta_t = \eta_{\max}\cdot\frac{t}{t_{\mathrm{warm}}}
\]
Gradual ramp-up prevents large gradient updates from random initialization from destabilizing early training.

Phase 2 -- Cosine annealing decay (steps 1,000--100,000)
\[
    \eta_t = \eta_{\min}
             + \tfrac{1}{2}(\eta_{\max}-\eta_{\min})
               \!\left(1+\cos\!\left(\pi\,
               \frac{t-t_{\mathrm{warm}}}{T-t_{\mathrm{warm}}}\right)
               \right)
\]
The cosine envelope provides a smooth, monotone decay, avoiding abrupt rate changes while continuing slow refinement at $\eta_{\min}=5\times10^{-5}$. To simulate an effective batch size of 1,024 on hardware limited to micro-batches of 32, gradients are accumulated over 32 steps before each optimizer update. Automatic mixed precision (AMP) with bfloat16 (float16 fallback) halves memory consumption and increases throughput. A \texttt{GradScaler} prevents underflow in float16 gradient computations. Memory-mapped binary files (NumPy \texttt{uint16} \texttt{memmap}) allow random-access sampling across the full corpus without loading it into RAM. Each step samples a random batch of context windows and constructs next-token prediction targets by shifting by one position. Every 500 training iterations, the model is placed in evaluation mode, and cross-entropy loss is averaged over 500 random validation batches. The checkpoint achieving the lowest validation loss is retained for final evaluation. Training was conducted on cloud instances (Tesla T4, P100) via Kaggle, Colab, and Lightning~AI. The full 100,000-iteration run required approximately $\sim$12 hours of wall-clock time and a peak VRAM footprint of $\sim$12\,GB. The mBERT and MuRIL continual pre-training experiments were conducted on a Tesla P100-PCIE-16\,GB (Kaggle), requiring approximately 2--3 hours each for 3 epochs over $\sim$27K training examples.

\section{Results}
\label{sec:results}



Three distinct phases characterize training. Rapid loss reduction from $\approx$7.5 to $\approx$2.0, driven by warmup initialization and steep gradient descent from 0--20,000 iterations. Sustained steady improvement as cosine decay refines learned representations from 20,000--50,000 iterations, and from 50,000--100,000 iterations, loss converges and stabilizes, resulting in a final training loss of $\approx$0.57, and a final validation loss of 0.5684.
\begin{figure}[h]
    \centering
    \small
    \includegraphics[width=0.5\linewidth]{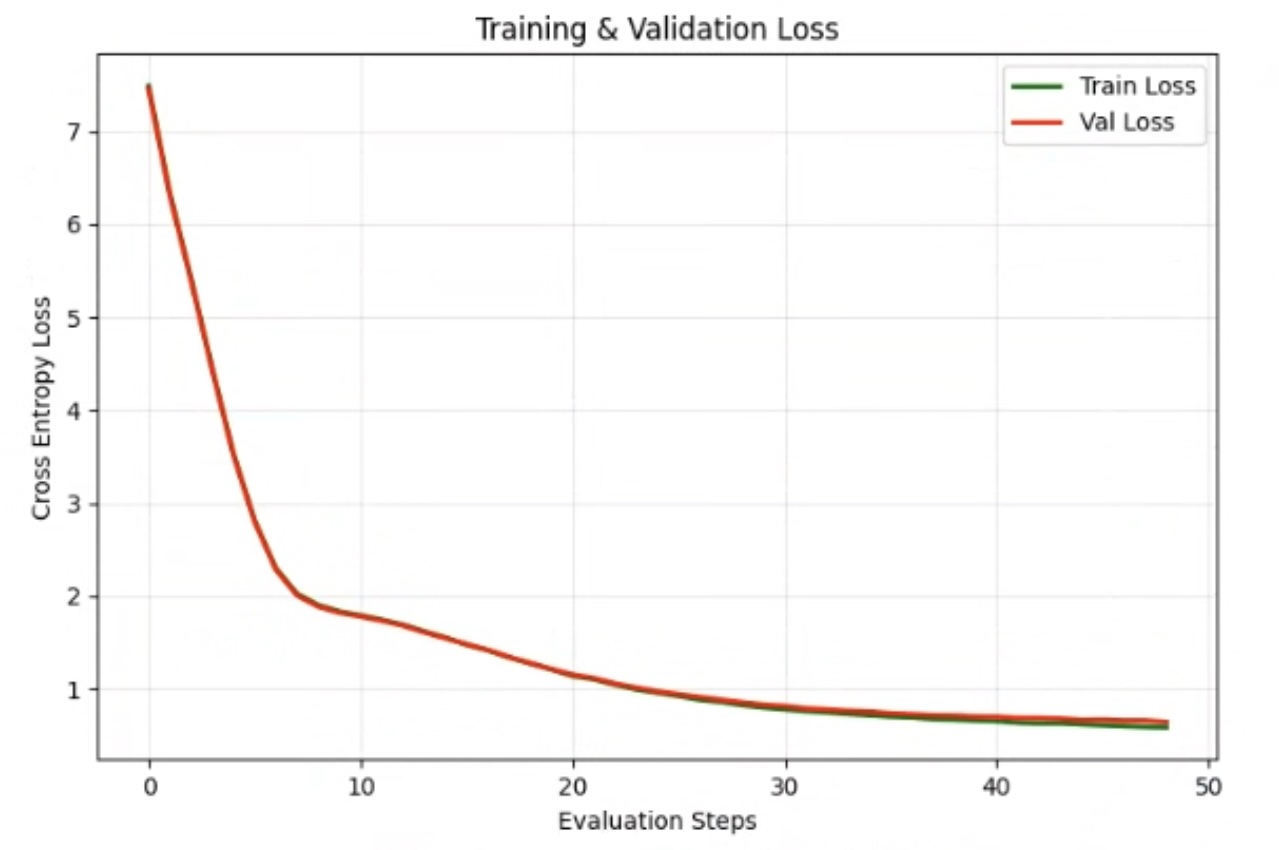}
    \caption{Training (green) and validation (red) cross-entropy loss vs.\ evaluation step (one step $=$ 500 iterations). Three phases: rapid descent (0--20k iterations), steady improvement (20--50k), and convergence near 0.57 (50--100k). The tight coupling of the two curves indicates strong generalization with negligible overfitting.}
    \label{fig:loss}
\end{figure}



\begin{figure}[htbp]
    \centering
    \includegraphics[width=0.55\linewidth]{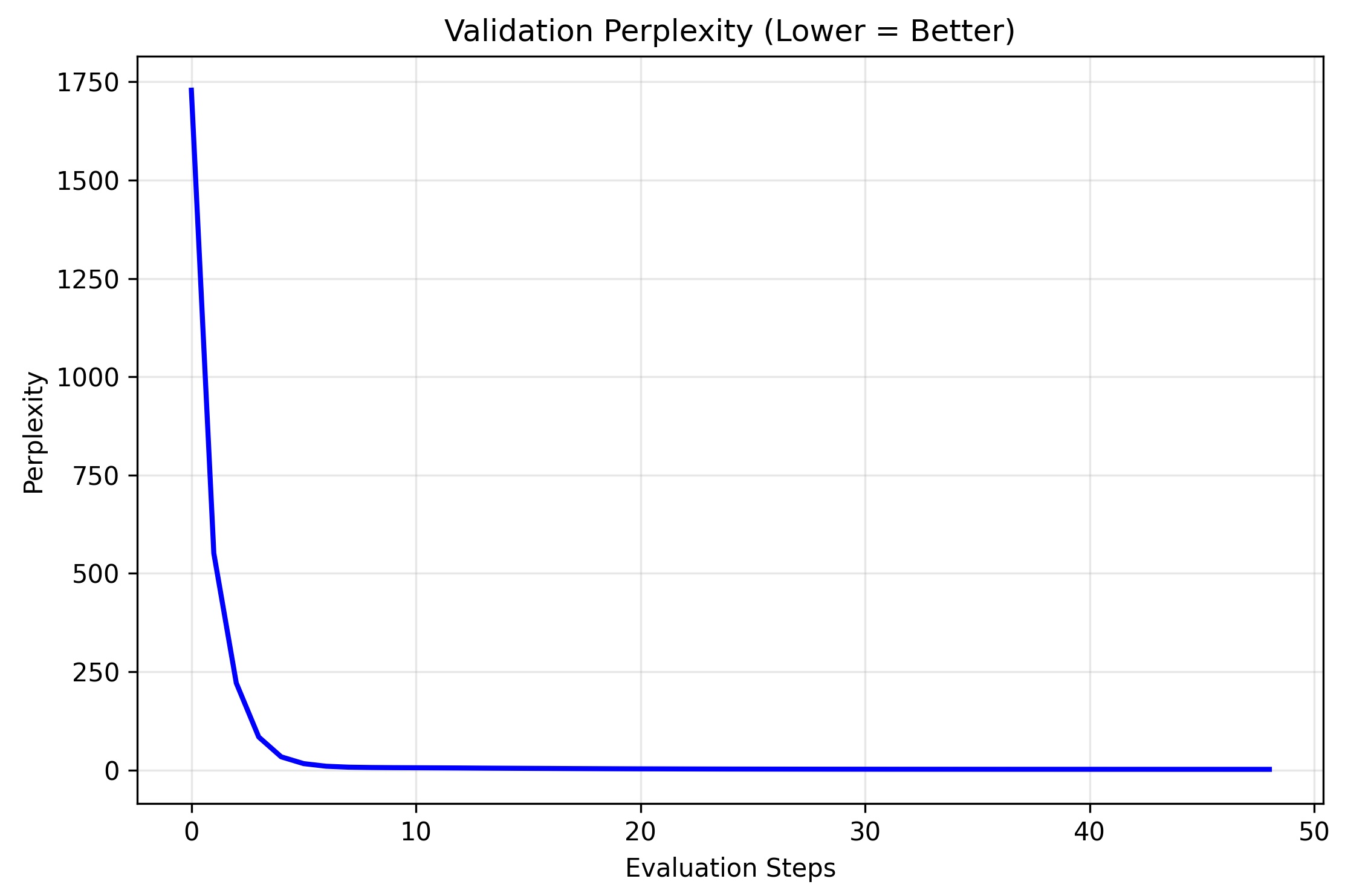}
    \caption{Validation perplexity (lower is better) over training. The reduction from $\sim$1,800 to 1.8 reflects deep adaptation to Nepali legal language patterns.}
    \label{fig:ppl}
\end{figure}

Perplexity $\mathrm{PPL} = \exp(\mathcal{L})$ provides an interpretable measure of per-token uncertainty (Figure~\ref{fig:ppl}). Starting from $\approx$1,800 at initialization, it falls sharply within the first 10,000 iterations before stabilizing at 1.8, meaning the model places the correct next token among its top-2 predictions on average.

\begin{figure}[htbp]
    \centering
    \includegraphics[width=0.55\linewidth]{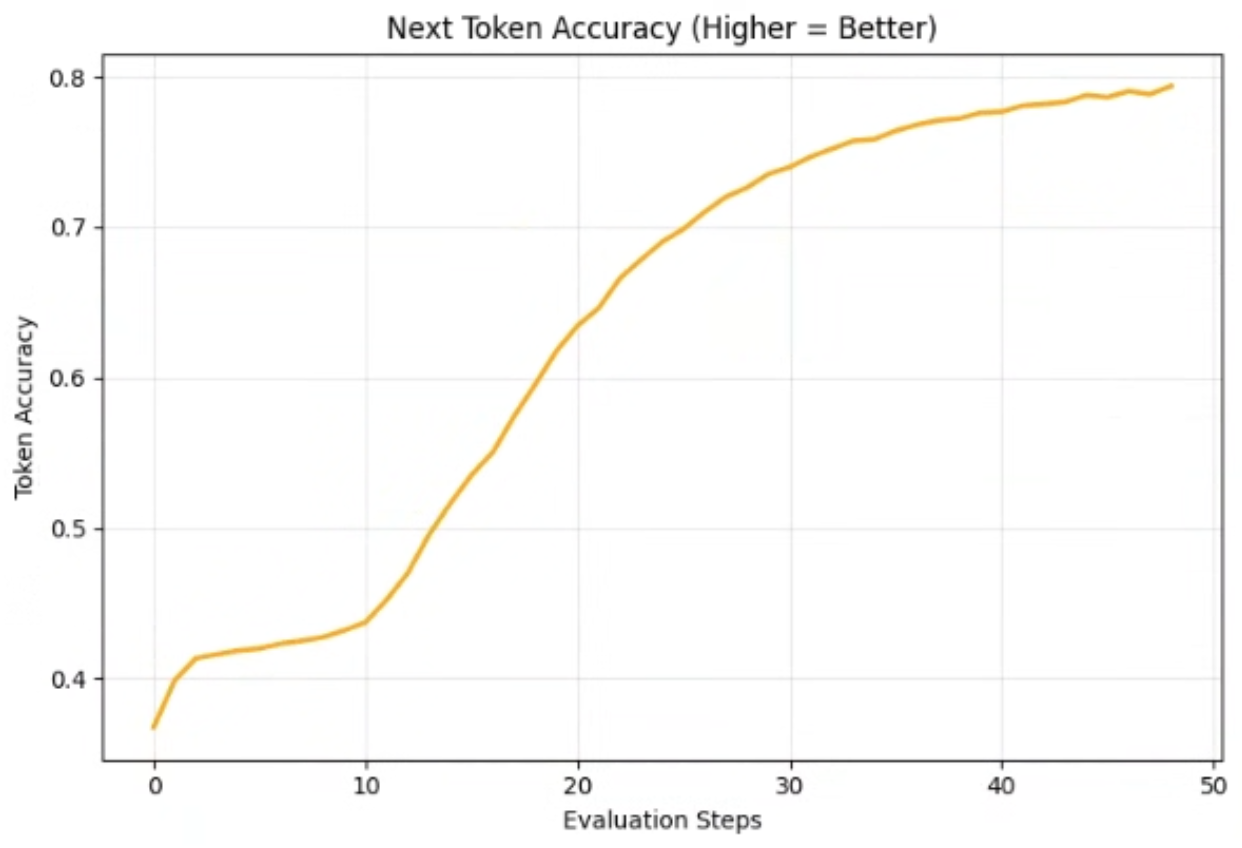}
    \caption{Top-1 next-token prediction accuracy on the validation set. Progression: $\approx$38\% (init) $\to$ $\approx$50\% (2k iter) $\to$ $\approx$75\% (10k iter) $\to$ 82.9\% (100k iter).}
    \label{fig:acc}
\end{figure}

Token prediction accuracy follows a characteristic learning curve, converging at 82.9\%, a strong result given the specialized vocabulary and complex syntax of Nepali legal text (Figure~\ref{fig:acc}).


\begin{figure}[htbp]
    \centering
    \includegraphics[width=0.55\linewidth]{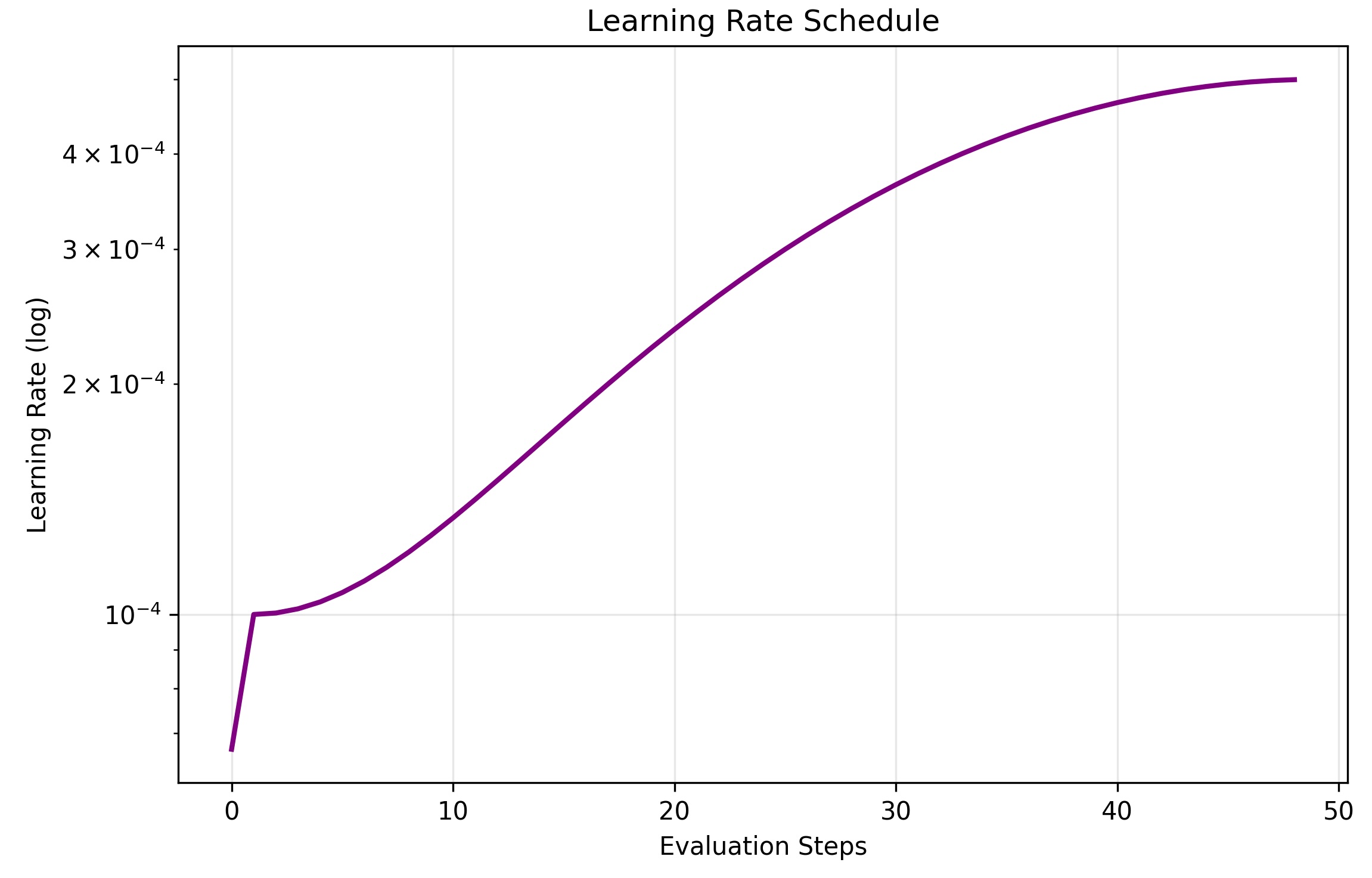}
    \caption{Two-phase learning rate schedule (log scale): linear warmup over steps 0--1,000, followed by cosine annealing decay over steps 1,000--100,000.}
    \label{fig:lr}
\end{figure}

\newpage
Table~\ref{tab:finalperf} summarizes the final metrics on the held-out validation split.

\begin{table}[h]
    \centering
    \small
    \caption{Final Model Performance on Held-Out Validation Set}
    \label{tab:finalperf}
    \begin{tabular}{@{}ll@{}}
        \toprule
        Metric & Value\\
        \midrule
        Validation loss (cross-entropy) & 0.5684 \\
        Perplexity                      & 1.8 \\
        Next-token accuracy (top-1)     & 82.9\% \\
        Training iterations             & 100,000\\
        Training duration               & $\sim$12 hrs\\
        Total parameters                & $\sim$30M \\
        Checkpoint size                 & $\sim$240\,MB\\
        \bottomrule
    \end{tabular}
\end{table}


Table~\ref{tab:baselines} compares NepLEGiT against all evaluated baselines on Nepali legal text, including the two encoder models subjected to continual masked-language-model pre-training.


\begin{table}[h]
    \centering
    \small
    \caption{Comparison Against Baseline Models on Nepali Legal Text}
    \label{tab:baselines}
    \begin{tabular}{@{}lrrr@{}}
        \toprule
        Model & Perplexity & Eval Loss & Accuracy \\
        \midrule
        Random baseline         & $\approx$50,257 & ---     & 0.002\% \\
        Unigram LM              & 500--1,000      & ---     & 10--15\% \\
        GPT-2 Small (zero-shot) & 40--60          & ---     & 45--55\%    \\
        \textbf{NepLEGiT (ours)}  & \textbf{1.8}    & \textbf{0.5684} & \textbf{82.9\%} \\
        \bottomrule
    \end{tabular}
\end{table}

NepLEGiT's perplexity of 1.8 represents a 22--33$\times$ improvement over GPT-2 Small despite using only 26\% of its parameters. Among the encoder models, mBERT (perplexity 2.35) substantially outperforms MuRIL (perplexity 6.07) after continual pre-training on the same corpus. These findings collectively confirm that targeted in-domain pre-training from scratch substantially outweighs scaling for specialized low-resource language tasks, consistent with the findings of \citet{eldan2023}.

\label{mbert_muril}

To complement NepLEGiT's generative decoder approach, we performed continual masked-language-model (MLM) pre-training of two established multilingual encoder models: mBERT (\texttt{bert-base-multilingual-cased}) and MuRIL (\texttt{google/muril-base-cased}) on the same Nepali legal corpus. Both models were fine-tuned for 3 epochs using the Hugging Face \texttt{Trainer} API on a Tesla P100-PCIE-16\,GB GPU, with the following shared configuration: batch size 2, gradient accumulation steps 16 (effective batch 32), learning rate $1\times10^{-5}$, cosine schedule, weight decay 0.01, max sequence length 256, and MLM probability 0.20.

\begin{table}[htbp]
    \centering
    \small
    \caption{NepLEGiT Models on Masked Language Modeling}
    \label{tab:mlm-model}
    \begin{tabular}{@{}lrr@{}}
        \toprule
        Model & Perplexity & Eval Loss\\
        \midrule
        \textbf{MuRIL (continual MLM) (ours)}   & 6.07            & 1.8026  \\
        \textbf{mBERT (continual MLM) (ours)}   & 2.35            & 0.8565  \\
        \bottomrule
    \end{tabular}
\end{table}

Table~\ref{tab:encoder_training} reports the validation loss trajectory at selected checkpoints.

\begin{table}[htbp]
    \centering
    \small
    \caption{Validation Loss During Continual MLM Pre-training (mBERT vs.\ MuRIL)}
    \label{tab:encoder_training}
    \begin{tabular}{@{}rcc@{}}
        \toprule
        Step & mBERT Val.\ Loss & MuRIL Val.\ Loss \\
        \midrule
        200  & 1.1983 & 2.2354 \\
        400  & 1.0751 & 2.1085 \\
        600  & 1.0040 & 2.0118 \\
        800  & 0.9548 & 1.9104 \\
        1000 & 0.9091 & 1.8633 \\
        1200 & 0.9067 & 1.8713 \\
        1400 & 0.8747 & 1.8467 \\
        1600 & 0.8602 & 1.8319 \\
        1800 & 0.8677 & 1.8014 \\
        2000 & 0.8547 & 1.8160 \\
        2200 & 0.8457 & 1.8070 \\
        \midrule
        \textbf{Final} & \textbf{0.8565} & \textbf{1.8026} \\
        \bottomrule
    \end{tabular}
\end{table}

mBERT converges to a final eval loss of 0.8565 (perplexity 2.35), while MuRIL converges to 1.8026 (perplexity 6.07). mBERT is therefore the stronger encoder baseline on this domain, despite MuRIL's explicit Indic-language focus. It is important to note that these perplexity values arise from a masked-language-modelling (MLM) objective and are not directly comparable to NepLEGiT's causal language-modelling (CLM) perplexity; the encoder models predict only 20\% of randomly masked tokens, while NepLEGiT predicts every next token autoregressively. These models serve distinct downstream roles: mBERT and MuRIL are better suited for classification, NER, and extractive QA, while NepLEGiT targets generative legal text production.

\section{Discussion}
\label{sec:analysis}

\subsection{Domain-Specific Pre-training}

The baseline comparison (Table~\ref{tab:baselines}) provides clear evidence of domain specificity's value. GPT-2 Small, with $4\times$ more parameters, achieves only 45--55\% token accuracy versus NepLEGiT's 82.9\%. Two factors explain this gap: (i)~GPT-2's tokenizer and weights encode English distributional statistics that poorly match Nepali Unicode character sequences; and (ii)~the domain shift from general web text to formal legal Nepali spans both language and register simultaneously. Targeted pre-training collapses both gaps at once.

\subsection{mBERT vs.\ MuRIL on Nepali Legal Text}
\label{mbertvsmuril}

A notable result of the encoder experiments is that mBERT outperforms MuRIL on the Nepali legal corpus despite the latter being explicitly pre-trained on Indic languages including Nepali \cite{khanuja2021}. We attribute this to two factors. First, MuRIL's pre-training emphasizes transliterated and code-mixed Indic text, whereas Nepali legal language is formal, monolingual, and highly Sanskritized, a register that differs substantially from conversational or transliterated usage. Second, mBERT's larger and more diverse multilingual pre-training may provide a more general-purpose linguistic scaffold that adapts more readily to a new domain via continual pre-training. This finding has practical implications: for Nepali legal NLP tasks requiring an encoder (e.g., named-entity recognition, document classification), mBERT is the stronger off-the-shelf starting point.

\subsection{Training Stability}

The near-identical train/validation loss curves (Figure~\ref{fig:loss}) over 100,000 iterations indicate the model does not overfit to the training split. Three design choices are responsible: dropout ($p=0.1$) at each sub-layer, weight decay ($\lambda=0.1$) via AdamW, and a model capacity (30M parameters) that is relatively small relative to the training corpus (4M tokens), leaving the model slightly underfitted --- which is preferable for a pre-training stage to be followed by fine-tuning.

\subsection{Perplexity Interpretation}

A perplexity of 1.8 on domain-specific text is notably low. GPT-2 (1.5B parameters) achieves perplexity around 18--25 on general English text \cite{radford2019}. The very low perplexity here reflects the high degree of formulaic repetition in legal Nepali; statutes reuse phraseological templates extensively, which a domain-specific model captures far more efficiently than a general-purpose one. Caution is warranted: low next-token perplexity does not directly translate to high downstream task performance (QA, summarization), which will require instruction fine-tuning.

\subsection{Tokenization Limitations}

The GPT-2 BPE tokenizer segments Nepali Devanagari suboptimally, producing longer token sequences than a native Nepali tokenizer would. This inflates effective sequence lengths and reduces contextual coverage within the 128-token window. We estimate that a custom Nepali BPE vocabulary (32K--50K tokens) trained on the legal corpus would reduce token counts by 20--35\%, enabling richer context at equivalent compute.

\subsection{Ethical Considerations}

NepLEGiT \emph{cannot replace qualified legal professionals}. Its outputs may contain hallucinated or factually incorrect legal information; all AI-generated legal content must be verified by licensed practitioners. The training corpus may encode historical biases present in Nepal's legal documentation (e.g., gender-asymmetric precedents). 




\section{Conclusion}
\label{sec:conclusion}

We have presented NepLEGiT from scratch, a language model pre-trained exclusively on Nepali legal text. Starting from a curated corpus of $\sim$4M tokens, we trained a $\sim$30M-parameter GPT-decoder SLM from scratch using mixed-precision arithmetic, gradient accumulation, and AdamW with warmup cosine-decay scheduling. The model achieves a validation perplexity of 1.8 and a next-token accuracy of 82.9\%, outperforming a zero-shot GPT-2 Small baseline by a factor of 22--33$\times$ in perplexity despite using only 26\% of its parameters. As a complementary encoder-side investigation, we performed continual pre-training and evaluated continual masked-language-model pre-training of mBERT and MuRIL on the same corpus. mBERT achieves an MLM perplexity of 2.35 (eval loss 0.8565), outperforming MuRIL (perplexity 6.07), establishing a strong encoder baseline for downstream understanding tasks.

NepLEGiT establishes a baseline for Nepali legal NLP and provides a fully documented, replicable methodology applicable to other low-resource, non-Western legal systems. The pre-trained model is positioned as the foundation for a broader legal AI ecosystem in Nepal: continual pre-training on multilingual backbones, supervised instruction fine-tuning, RAG-augmented long-document reasoning, and a publicly accessible legal information service for Nepali citizens, legal professionals, and institutions. By demonstrating that focused domain pre-training at modest scale decisively outperforms general-purpose models for this specialized task, NepLEGiT advances the case that low-resource legal AI is tractable and consequential for equitable access to justice. We prefer a custom Nepal legal BPE tokenizer, continual pre-training on multilingual decoder-based models, supervised fine-tuning on domain-specific downstream tasks, and RAG-based informal retrieval systems as future work.

\bibliographystyle{splncs04}

\end{document}